\documentclass[12pt]{article}
\usepackage[utf8]{inputenc}
\usepackage[margin=1in]{geometry}
\usepackage{natbib}
\usepackage{enumitem}
\usepackage{amsmath,amssymb}
\usepackage{makecell}
\usepackage{adjustbox}
\usepackage{graphicx}
\usepackage{float}
\usepackage{xcolor}
\usepackage{booktabs}
\usepackage{authblk}
\usepackage{setspace}
\usepackage{comment}
\usepackage{units}
\usepackage[ruled,vlined]{algorithm2e}
\usepackage[citecolor=blue,colorlinks=true,linkcolor=blue]{hyperref}
\usepackage[hyperref]{ntheorem}

\DeclareMathOperator*{\argmax}{arg\,max}

\title{Missing Velocity in Dynamic Obstacle Avoidance based on Deep Reinforcement Learning}
\author[*]{Fabian Hart}
\author[*]{Martin Waltz}
\author[*]{Ostap Okhrin}
\affil[*]{Institute of Transportation Economics, Technische Universität Dresden, Germany}
\begin{document}
\sloppy
\definecolor{alizarin}{rgb}{0.82, 0.1, 0.26}
\maketitle

\begin{abstract}
    We introduce a novel approach to dynamic obstacle avoidance based on Deep Reinforcement Learning by defining a traffic type independent environment with variable complexity. Filling a gap in the current literature, we thoroughly investigate the effect of missing velocity information on an agent's performance in obstacle avoidance tasks. This is a crucial issue in practice since several sensors yield only positional information of objects or vehicles. We evaluate frequently-applied approaches in scenarios of partial observability, namely the incorporation of recurrency in the deep neural networks and simple frame-stacking. For our analysis, we rely on state-of-the-art model-free deep RL algorithms. The lack of velocity information is found to significantly impact the performance of an agent. Both approaches - recurrency and frame-stacking - cannot consistently replace missing velocity information in the observation space. However, in simplified scenarios, they can significantly boost performance and stabilize the overall training procedure.
\end{abstract}

\newpage
\section{Introduction}
The problem of deriving a collision-free path for an agent moving among dynamic obstacles is a widely studied area and has applications in many fields of automated transportation systems, such as self-driving cars \citep{OAinAutonomousDriving2008}, unmanned aerial vehicles \citep{OAinUAV2010}, and service robots \citep{OAinServiceRobotics2016}. However, \cite{OAnpHard1987} showed that in a simple obstacle avoidance (OA) case, where a 2D holonomic robot faces dynamic polygon obstacles with constant velocities, the problem is NP-Hard. Velocity Obstacles \citep{VO1998} is one of several algorithmic solutions that have been developed for the problem of dynamic OA and has been widely applied on vehicles for collision prevention, e.g., wheeled robots \citep{VOonWheeledRobots2009}, unmanned aerial vehicles \citep{VOonUSVs2016}, and unmanned surface vehicles \citep{VOonUSV}.

At present, the advances in machine learning methods, particularly in Reinforcement Learning (RL, \cite{sutton2018reinforcement}), provide a new possibility for navigation in dynamic environments. Especially Deep RL, which uses deep neural networks \citep{goodfellow2016deep} as function approximators, has already shown remarkable achievements, e.g., by learning to play Atari video games from pixels \citep{mnih2015human} or by mastering the game of Go \citep{Silver2017alphaGo}. These methods have also been successfully applied to the domain of obstacle avoidance: \cite{DUGULEANA2016} and \cite{Cimurs2020} use RL to compute collision-free trajectories for mobile robots in real-life environments; \cite{Wang2019} and \cite{Roghair2021} train agents to allow unmanned aerial vehicles to navigate in complex environments; \cite{Chen2019} and \cite{XU2020} propose collision avoidance algorithms for underactuated unmanned surface vehicles using RL. Another strand of literature uses vision- or rangefinder-based information from the environment and are primarily based on convolutional neural networks to extract features about surrounding obstacles, see \citep{xie2017monocular}, \citep{Cimurs2020}, or \citep{Wang2018}. Furthermore, various studies \citep{Bhopale2019, XU2020, Yan2021} focus on RL-based OA algorithms that directly use hand-shaped features about surrounding obstacles, for example, positions or headings. Common practice is also the inclusion of velocity information about obstacles into these features, although in real applications only relative distances to obstacles can be extracted from many common sensors, e.g., camera image data. However, there is to the best of our knowledge no comprehensive comparison of how severe this velocity information loss affects the performance of the used algorithms. To stress: all aforementioned studies either do use or do not use velocity information explicitly or implicitly.

This motivated us to thoroughly compare both approaches and analyze the resulting performances of collision prevention in a generic OA environment. Furthermore, common strategies to combat velocity information deficiencies and to improve the trajectory anticipation capabilities of the agent include: 1) the use of a time-series of past environmental information; and 2) the incorporation of recurrency into the neural network structure \citep{Altche2017}. Therefore, we additionally analyzed whether recurrent layers can boost the overall performance if velocity information about obstacles is missing. Summarizing, the main contributions of our work are as follows:
\begin{itemize}
    \item Introduction of a novel approach to dynamic OA based on Deep RL including the definition of a generic environment with variable complexities.
    \item Analysis of a sensory-motivated reduced observation space in which information about velocities is not available.
    \item A comprehensive comparison of state-of-the-art model-free RL algorithms for continuous action spaces with and without recurrency in the deep neural network structure.
\end{itemize}
Based on the previous research, we formulate and test the following hypotheses:
\begin{itemize}[leftmargin=*]
    \item[] \emph{Hypothesis 1}: One can use recurrent layers in the function approximation to reconstruct missing velocity information solely from positional information in an obstacle avoidance task.
    \item[] \emph{Hypothesis 2}: Alternatively, one could simply use frame-stacking to reconstruct missing velocity information.
\end{itemize}
This work is structured as follows: In Section \ref{sec:RLmethodology}, we give a detailed overview of the RL basics and we provide information about the used RL algorithms. In Section \ref{sec:Approach}, we define the OA environment variants to test our hypotheses, followed by the results of the RL training in Section \ref{sec:results}. The results are discussed in Section \ref{sec:Discussion}. Section \ref{sec:Conclusion} concludes.

\section{Reinforcement Learning Methodology}
\label{sec:RLmethodology}
\subsection{Basics}\label{subsec:RL_basics}
RL aims at solving sequential decision tasks in which an agent interacts with an environment under the objective to maximize the received reward \citep{sutton2018reinforcement}. Formally, we consider Markov Decision Processes (MDP) consisting of a state space $\mathcal{S}$, an action space $\mathcal{A}$, an initial state distribution $T_0: \mathcal{S} \rightarrow [0,1]$, a state transition probability distribution $\mathcal{P}: \mathcal{S} \times \mathcal{A} \times \mathcal{S} \rightarrow [0,1]$, a reward function $\mathcal{R}: \mathcal{S} \times \mathcal{A} \rightarrow \mathbb{R}$, and a discount factor $\gamma \in [0,1]$. At each time step $t$, the agent receives a state information $S_t \in \mathcal{S}$, selects an action $A_t \in \mathcal{A}$, gets a reward $R_{t+1}$, and transitions based on the environmental dynamics $\mathcal{P}$ to the next state $S_{t+1} \in \mathcal{S}$. Furthermore, we consider Partially Observable Markov Decision Processes (POMDP, \cite{kaelbling1998planning}), which generalize the MDP by introducing two additional components: the observation space $\mathcal{O}$ and the observation function $\mathcal{Z}: \mathcal{S} \times \mathcal{A} \times \mathcal{O} \rightarrow [0,1]$. In a POMDP, the agent does not receive the new state $S_{t+1}$ directly, but instead an observation $O_{t+1} \in \mathcal{O}$, which is generated with probability $P(O_{t+1} | S_{t+1}, A_t)$ by the observation function $\mathcal{Z}$. Consequently, a POMDP is a Hidden Markov Model with actions and the observations are used for learning. In the following, we use capital notation, e.g., $S_t$, to indicate random variables and small notation, e.g., $s_t$ or $s$, to describe realizations.

Objective of the agent in the MDP scenario is to learn a policy $\pi: \mathcal{S} \times \mathcal{A} \rightarrow [0,1]$, a mapping from states to probability distributions over actions, that maximizes the expected return, which is the expected discounted cumulative reward, from the start state: $E_{\pi}\left[ \sum_{k=0}^{\infty} \gamma^k R_{k+1}\right | S_0]$. Common practice is the definition of action value functions $Q^{\pi}(s,a)$, which are the expected return when starting in state $s$, taking action $a$, and following policy $\pi$ afterward: $Q^{\pi}(s,a) = E_{\pi}\left[ \sum_{k=0}^{\infty} \gamma^k R_{t+k+1} | S_t = s, A_t = a \right]$. Crucially, in an MDP there is always a deterministic optimal policy $\pi^*(s) = \argmax_{a \in \mathcal{A}} Q^*(s,a)$, that is connected with an optimal action-value function $Q^*(s,a) = \max_{\pi} Q^{\pi}(s,a)$. To learn $Q^*(s,a)$, a recursive relationship termed Bellman optimality equation \citep{bellman1954theory} is frequently used:
\begin{equation}\label{eq:Bellman_opt_eq}
    Q^*(s,a) = \mathcal{R}(s,a) + \gamma \sum_{s' \in \mathcal{S}} \mathcal{P}_{sa}^{s'} \max_{a' \in \mathcal{A}} Q^*(s',a').
\end{equation}

The popular Q-Learning algorithm \citep{watkins1992q} translates (\ref{eq:Bellman_opt_eq}) into a sample-based update procedure. The Q-values are approximated by tabular representations, which store a particular value for each $(s,a)$-pair. However, this approach is not feasible for continuous state spaces, which is why more complex representations like deep neural networks are used to approximate the Q-values. This serves as a basis for the Deep Q-Network (DQN, \cite{mnih2015human}), which is a fundamental approach to combine Q-Learning with function approximation.~Having a function $Q^{\omega}(s,a)$ with parameter vector $\omega$, the training is realized by gradient descend:
\begin{equation}\label{eq:DQN_grad_descent}
    \omega \leftarrow \omega + \alpha \left\{y-Q^{\omega}(s,a) \right\} \nabla_{\omega} Q^{\omega}(s,a), 
\end{equation}
with reward $r$, target $y = r + \gamma \max_{a' \in \mathcal{A}} Q^{\omega'}(s',a')$, and learning rate $\alpha$. $Q^{\omega'}(s,a)$ is referred to as the target network and can greatly stabilize the training process. It is a time-delayed copy of the original network with parameter $\omega$. Furthermore, DQN uses experience replay, in which past transitions are sampled randomly (or with more sophisticated strategies like \cite{schaul2015prioritized}) to perform gradient descent steps. However, DQN is restricted to discrete action spaces $\mathcal{A}$ since it involves calculating the maximum over all possible actions. Our application case involves continuous actions spaces, which is the reason we use the state-of-the-art TD3 algorithm \citep{fujimoto2018addressing}. Its functionality is detailed in the following.\\

\subsection{Twin Delayed Deep Deterministic Policy Gradient (TD3)}

The TD3 is an extension of the Deep Deterministic Policy Gradient (DDPG) algorithm of \cite{lillicrap2015continuous}. The DDPG is an off-policy, actor-critic algorithm that uses neural networks as function approximators. Importantly, it is based on a \emph{deterministic} policy $\mu^{\theta}: \mathcal{S} \rightarrow \mathcal{A}$ with parameter vector $\theta$. In this setup $\mu^{\theta}$ takes the role of the actor and approximates the maximum operation in the target computation. The second component of the framework is the critic function $Q^{\omega}(s,a)$, which approximates the action-values as in the DQN and is updated by gradient descent. In this context, the critic will be used to evaluate the actions made by the actor. More precisely, we consider the performance objective based on the deterministic policy: $J(\mu^{\theta}) = E_{\mu^{\theta}}\left[ \sum_{k=0}^{\infty} \gamma^k R_{k+1}\right | S_0]$. \cite{silver2014deterministic} proved the \emph{Deterministic Policy Gradient Theorem}, which yields the gradient of the performance measure with respect to $\theta$:
\begin{equation}\label{eq:deter_pol_grad_theorem}
    \nabla_{\theta}J(\mu^{\theta}) \approx E_{s \sim \rho^{\mu}} \left\{\nabla_{\theta} \mu^{\theta}(s) \nabla_a Q^{\omega}(s,a)|_{a=\mu^{\theta}(s)} \right\},
\end{equation}
where $\rho^{\mu}$ is the discounted state visitation distribution. This gradient can be used to train the actor via gradient ascent, so that both actor and critic are updated iteratively. Furthermore, \cite{lillicrap2015continuous} proposed to also use experience replay and target networks. However, a soft-update of the target networks for both actor and critic is applied, which constrains the update targets to change slowly and yields a further stabilized training procedure. Denoting $\tau$ as the soft target update rate, $\theta'$ and $\omega'$ the parameter sets of the target actor and critic, respectively, the update is:
\vspace{-0.5cm}
\begin{align}\label{eq:DDPG_soft_tgt_up}
    \omega' &= \tau \omega + (1-\tau) \omega', \nonumber \\ 
    \theta' &= \tau \theta + (1-\tau) \theta'.
\end{align}
Exploration is performed by perturbing the action of the actor with additional random noise. However, \cite{fujimoto2018addressing} introduced three modifications of the original DDPG to receive a state-of-the-art model-free algorithm. First, the TD3 uses the minimum of two critics $Q^{\omega_1}(s,a)$ and $Q^{\omega_2}(s,a)$ to combat the overestimation issue in the critic update. Second, the variance of the critic update is reduced by introducing target policy smoothing. Consequently, while the critic target in DDPG was $y = r + \gamma Q^{\omega'}\left\{s', \mu^{\theta'}(s') \right\}$, the TD3 uses $y = r + \gamma \min_{i=1,2} Q^{\omega'_i}\left\{s', \mu^{\theta'}(s') + \Tilde{\epsilon}\right\}$ with $\Tilde{\epsilon} \sim \text{clip}\{\mathcal{N}(0, \Tilde{\sigma}),-c,c\}$ for some $c > 0$, and normal distribution $\mathcal{N}$ with standard deviation $\Tilde{\sigma}$. Third, instead of performing policy and target updates at every step, the TD3 typically performs them only every $d = 2$ steps, which was shown to yield improved performance. The complete algorithm is detailed in Appendix \ref{appendix:TD3_algo}.

\subsection{Long-Short-Term-Memory (LSTM) based TD3}
As described in Section \ref{subsec:RL_basics}, only observations $o_t$ rather than full states $s_t$ are available in the POMDP case. One popular approach to handle this scenario is the construction of belief states, which are distributions over the real states the agent might be in, given the observation so far. However, this requires a model of the environment and is computationally demanding \citep{heess2015memory}. An alternative approach might be to stack past observations together (see \cite{mnih2015human}) and use this as input for the network. This frame-stacking (FS) technique will also be investigated in Section \ref{sec:results}, where we equip the TD3 algorithm with past observations and refer to it as TD3-FS. However, it is not immediately obvious which information will be of relevance later on, and all past observations are equally weighted when simply expanding the input vector. Finally, a further approach is to incorporate recurrency into the function approximators of model-free algorithms, which was shown to be capable of strong performances \citep{ni2021recurrent}. The recurrency enriches the agent's decision making by extracting information of past observations, potentially yielding an improved ability to solve problems without access to the complete state vector. Concretely, \cite{meng2021memory} proposed an extension of the TD3 called LSTM-TD3, which adds LSTM layers \citep{hochreiter1997long} to actor and critic of the TD3. The resulting algorithm showed impressive results on several benchmark tasks from the continuous action domain. We adapt it as our memory-based model-free competitor to the TD3.\\

\begin{figure}[!ht]
    \centering
    \includegraphics[width=0.65\linewidth]{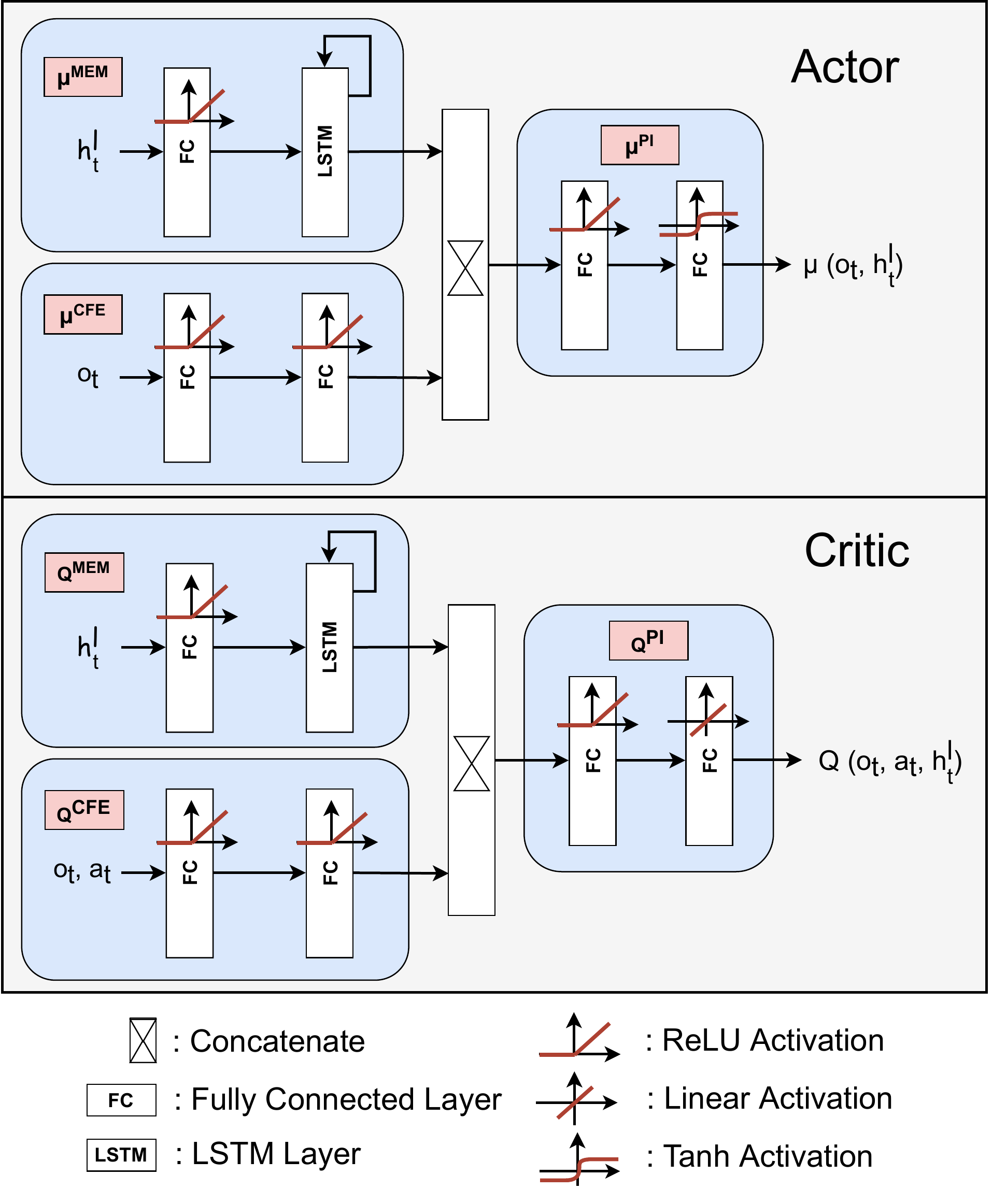}
    \caption{Illustration of the implemented LSTM-TD3 network architecture, adapted from \cite{meng2021memory}. MEM abbreviates memory extraction, CFE is current feature extraction, and PI refers to perception integration. Note that with \emph{one layer}, we mean \emph{one weight connection matrix}, as it is specified when implementing the architecture in deep learning frameworks like PyTorch \citep{paszke2019pytorch} or Tensorflow \citep{abadi2016tensorflow}. This illustration format is chosen since it immediately enables reproducability.}
    \label{fig:ActorCritic_LSTM}
\end{figure}

In the following, we use the notation $o_t$ instead of $s_t$ since the LSTM-TD3 was developed to tackle POMDP scenarios. However, in the dynamic OA scenario detailed in Section \ref{sec:Approach}, we test all approaches (TD3, TD3-FS, LSTM-TD3) with full state and reduced observation input, respectively. To describe the functionality of LSTM-TD3, we define the past history $h_{t}^{l}$ of length $l$ at time step $t$ as:
\begin{equation}
h_t^l = \begin{cases} 
      o_{t-l}, \ldots, o_{t-1} & \text{if \hspace{0.125cm}} l, t \geq 1. \\
      o_0 & \text{else}. 
   \end{cases}
\end{equation}
$o_0$ is a zero-valued dummy observation of the same dimension as a regular observation. Note that the defintion of $h_{t}^{l}$ slightly differs from \cite{meng2021memory} since we do not include past actions in the history. Furthermore, we set $l = 2$ throughout the paper, because, from a physical perspective, velocity and acceleration of an obstacle can be estimated based on its current and two last positions. The algorithm disassembles both actor and critic into different sub-components. Precisely, there is a memory extraction (MEM) part in the function approximators, $Q^{mem}$ and $\mu^{mem}$, respectively, that processes the history. In parallel, the current feature extraction (CFE) components $Q^{cfe}$ and $\mu^{cfe}$ process the observation of the current step $o_t$. Finally, the output of both MEM and CFE are concatenated and fed into the perception integration (PI) components $Q^{pi}$ and $\mu^{pi}$. These aggregate the extracted pieces of information and yield the final result. The complete network design of our LSTM-TD3 implementation is illustrated in Figure \ref{fig:ActorCritic_LSTM} and formalized as follows:
\begin{align}
    Q(o_t, a_t, h_{t}^{l}) &= Q^{pi}\left\{Q^{me}(h_{t}^{l}) \bowtie Q^{cfe}(o_t, a_t) \right\},\\
    \mu(o_t, h_{t}^{l}) &= \mu^{pi}\left\{\mu^{me}(h_{t}^{l}) \bowtie \mu^{cfe}(o_t) \right\},
\end{align}
where $\bowtie$ is the concatenation operator. The remaining optimization and training process follows the one of the TD3. Algorithm \ref{algo:LSTM_TD3} summarizes the procedure.

\begin{algorithm}
\setstretch{1.10}
\SetAlgoLined
 Randomly initialize critics $Q^{\omega_1}, Q^{\omega_2}$ and actor $\mu^{\theta}$\\
 Initialize target critics $Q^{\omega'_1}, Q^{\omega'_2}$ and target actor $\mu^{\theta'}$ with $\omega'_1 \leftarrow \omega_1$, $\omega'_2 \leftarrow \omega_2$, $\theta^{'} \leftarrow \theta$\\
 Initialize replay buffer $\mathcal{D}$\\
 Receive initial observation $o_1$ from environment, initialize history $h_{1}^{l} = \mathbf{0}$ \\
 \For{t = 1,T}{
 \emph{Acting}\\
 Select action with exploration noise: $a_t = \mu^{\theta}(o_t, h_{t}^{l}) + \epsilon$, \quad $\epsilon \sim \mathcal{N}(0, \sigma)$\\
 Execute $a_t$, receive reward $r_{t+1}$, new observation $o_{t+1}$, and done flag $d_t$\\
 Store transition $(o_t, a_t, r_{t+1}, o_{t+1}, d_t)$ to $\mathcal{D}$\\
 \medskip
 \emph{Learning}\\
  Sample random mini-batch of transitions with their corresponding histories $\left(h_{i}^{l}, o_{i}, a_{i}, r_{i+1}, o_{i+1}, d_{i}\right)_{i=1}^{N} $ from $\mathcal{D}$\\
  Calculate targets:\\
  \vspace{-1cm}
  \begin{align*}
      \Tilde{a}_{i+1} &= \mu^{\theta'}(o_{i+1}, h_{i+1}^{l}) + \Tilde{\epsilon}, \quad \Tilde{\epsilon} \sim \text{clip}\{\mathcal{N}(0, \Tilde{\sigma}),-c,c\},\\
      y_i &= r_{i+1} + \gamma (1-d_{i}) \min_{j=1,2} Q^{\omega'_j}(o_{i+1},\Tilde{a}_{i+1}, h_{i+1}^{l}).
  \end{align*} \\
  \vspace{-0.5cm}
 Update critics: $\omega_j \leftarrow \min_{\omega_j}N^{-1} \sum_i \left\{y_i - Q^{\omega_j}(o_{i}, a_{i}, h_{i}^{l})\right\}^2$\\
 \If{$t \mod d$}{
 Update actor: $\theta \leftarrow \max_{\theta} N^{-1} \sum_{i} Q^{\omega_1}\left\{o_i, \mu^{\theta}(o_{i}, h_{i}^{l}), h_{i}^{l}\right\}$\\
  Update target networks via (\ref{eq:DDPG_soft_tgt_up})
 }
 \medskip
\emph{End of episode handling}\\
    \uIf{$d_t$}{
    Reset environment to get initial observation $o_{t+1}$\\
    Reset history $h_{t+1}^{l} = \mathbf{0}$\\
    }\Else{
    $h_{t+1}^{l} = (h_{t}^{l} - o_{t-l}) \cup o_t$
    }
}
 \caption{LSTM-TD3 algorithm following \cite{meng2021memory}.}
 \label{algo:LSTM_TD3}
\end{algorithm}

\subsection{Implementation and Initial Example}\label{subsec:implementation}
\begin{figure}
    \centering
    \includegraphics[width=\linewidth]{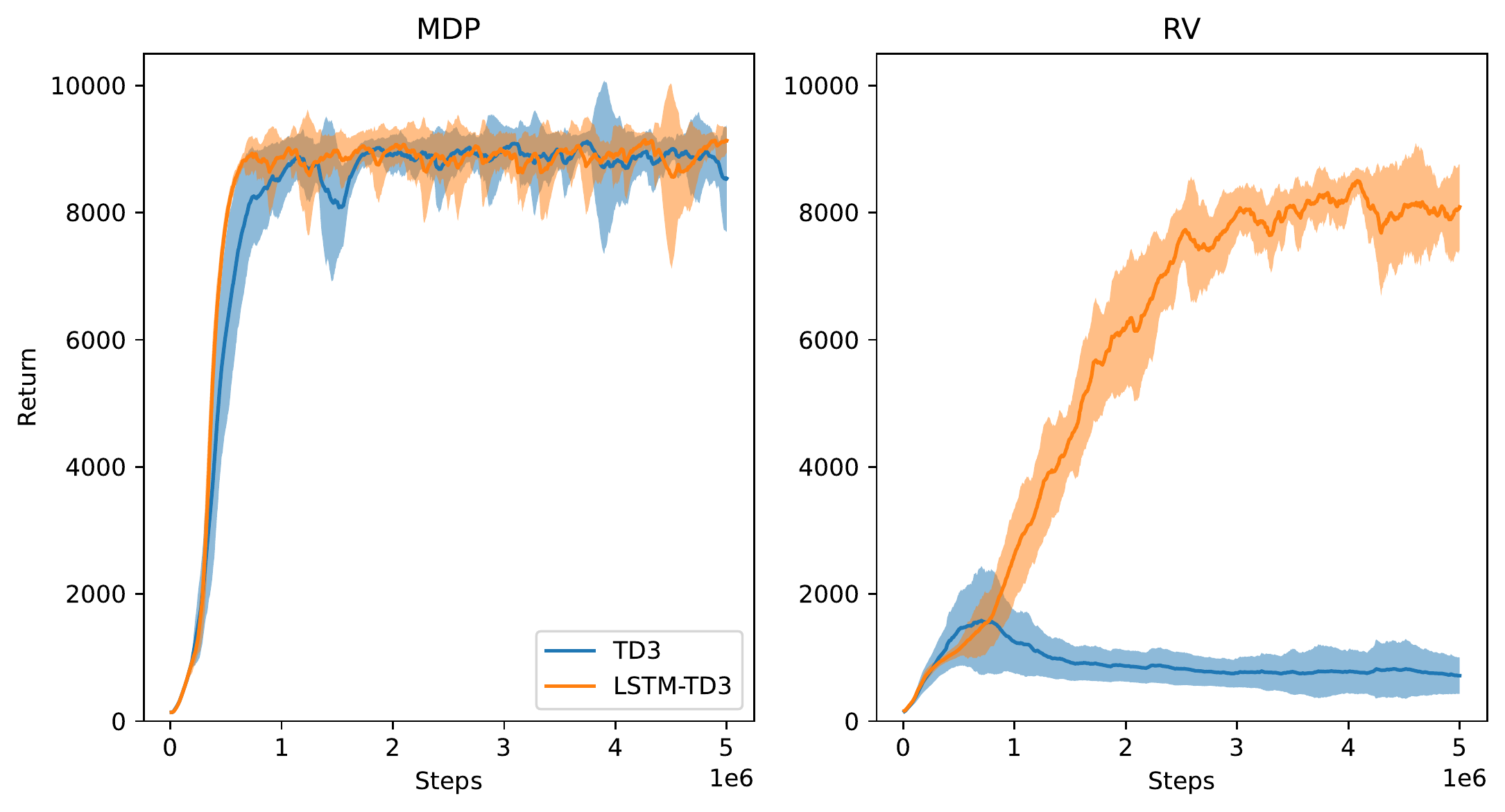}
    \caption{Performance comparison of TD3 with LSTM-TD3 for the environment  \texttt{InvertedDoublePendulumPyBulletEnv-v0}. Results are averaged over 10 independent runs. The shaded area are two standard deviations over the runs.}
    \label{fig:InvDouPendulum}
\end{figure}

All algorithms and environments shown in this paper are implemented in Python while using the deep learning framework PyTorch \citep{paszke2019pytorch}. Optimization is performed with Adam \citep{kingma2014adam}. The complete list of hyperparameters is given in Appendix \ref{appendix:TD3_algo}, while we specify the network structure of the TD3 as in the original proposal of \cite{fujimoto2018addressing}. To initially validate the performance potential of the LSTM-TD3 over the TD3 when POMDP cases are present, we select the \verb"InvertedDoublePendulumPyBulletEnv-v0" environment from PyBullet-Gym \citep{pybullet}. This is a classic continuous control problem. More precisely, we consider the MDP version with a fully observable state-space, and the POMDP version called Remove-Velocity (RV), in which velocity-related elements of the state-vector are not available. In theory, if velocities are important to master a given task, the LSTM-TD3 algorithm should have severe advantages over the TD3 due to its processing of past information. We train each algorithm for $5 \cdot 10^6$ time steps. Every $5\,000$ training steps, we average the return of 10 evaluation episodes, which are played with the current deterministic policy. The whole procedure is repeated for 10 different seeds and exponentially smoothed for clarity. Figure \ref{fig:InvDouPendulum} shows the results. We observe that both algorithms learn relatively fast and stable in the MDP case, and reach a similar level of final performance. However, the TD3 is clearly not able to master the given task in the RV scenario since it initially learns fast, but than drops back to a low level of performance. In contrary, the LSTM-TD3 has a longer and relatively stable learning performance, reaching a final level nearly comparable to the MDP case.

\section{Approach: Obstacle Avoidance}\label{sec:Approach}
\subsection{Problem Description}

To test our initial hypotheses, we propose two different obstacle avoidance environments, on which we thoroughly compare different RL algorithms. We distinguish between an MDP scenario, which includes the full state information, and an RV case, in which velocity information is not available. The main objective is to analyze the performance of the algorithms when hiding velocity information in the observation of the agent. We try to formulate general obstacle avoidance environments that do not dependent on a specific type of traffic. This leads to the following assumptions:
\begin{itemize}
	\item The agent, as well as the obstacles, are represented as point mass models.
	\item The agent's speed in the longitudinal direction is constant, while the lateral dynamics are controlled by the agent.
	\item The obstacle velocities are constant.
	\item The obstacles can be passed only in a predefined fashion, thus avoiding binary passing decisions where the agent may get stuck in between obstacles.
\end{itemize}
Under these assumptions and thinking of the obstacles as other traffic participants, we can consider our environment as a general representation of two-dimensional traffic with overtaking rules, for example, inland vessel traffic \citep{InlandVesselTraffic}. 

\begin{figure}[ht]
	\centering
	\includegraphics[width=\linewidth]{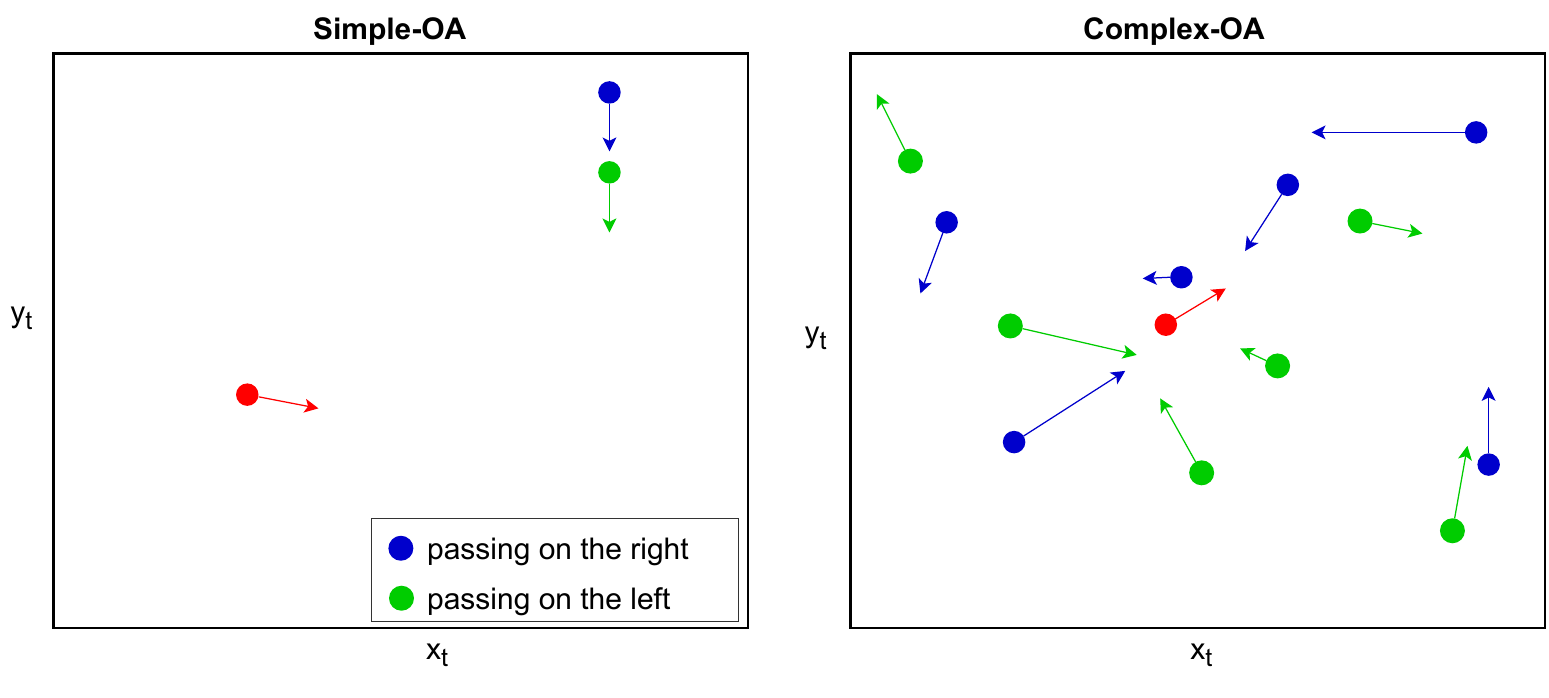}
	\caption{Environments Simple-OA and Complex-OA, where the agent is red, obstacles with passing rule 'right' are blue, and obstacles with passing rule 'left' are green. The arrows indicate direction and magnitude of velocity.}
	\label{fig:Envs_general}
\end{figure}

Both environments are depicted in Figure \ref{fig:Envs_general}. The first one, Simple-OA, is a reduced scenario with two obstacles and additional constraints that allow to isolate the problem of velocity reconstruction. The second environment, Complex-OA, follows the same principles but represents a more complex and realistic obstacle avoidance environment. In the following, we describe both environmental dynamics and the corresponding simulation procedures in detail.

\subsection{General Environment Definitions}

We consider a set of obstacles $\mathcal{M} = \{1, \ldots, N_{\rm obstacle}\}$, where $N_{\rm obstacle}$ is the total number of obstacles in the respective environment. For each time step $t$, we define $x_{t,\rm agent}$ and $x_{t,i}$ as the longitudinal position of agent and obstacle $i \in \mathcal{M}$, respectively, and $y_{t,\rm agent}$ and $y_{t,i}$ as the corresponding lateral positions. $\Dot{x}_{t,\rm agent}$ and $\Dot{x}_{t,i}$ denote the longitudinal speed, and $\Dot{y}_{t,\rm agent}$ and $\Dot{y}_{t,i}$ the lateral speed for agent and obstacle $i \in \mathcal{M}$, while $\Ddot{y}_{t,\rm agent}$ is the agent's lateral acceleration. Based on those definitions, the state at time step $t$ is defined as:
\begin{equation}
	s_{t} = \begin{pmatrix}
		\frac{\Ddot{y}_{t,\rm agent}}{a_{y,\rm max}}\\
		\frac{\Dot{y}_{t,\rm agent}}{v_{y,\rm max}}\\
		\frac{\Dot{x}_{t,\rm agent}-\Dot{x}_{t,i}}{v_{x,\rm max}}\\
		\frac{\Dot{y}_{t,\rm agent}-\Dot{y}_{t,i}}{v_{y,\rm max}}\\
		\frac{x_{t,\rm agent}-x_{t,i}}{x_{\rm scale}}\\
		\frac{y_{t,\rm agent}-y_{t,i}}{y_{\rm scale}}
	\end{pmatrix},
\end{equation}
where $a_{y,\rm max}$ defines the maximum lateral acceleration for the agent, $v_{x,\rm max}$ and $v_{y,\rm max}$ denote maximum lateral and longitudinal speeds, and $x_{\rm scale}$ and $y_{\rm scale}$ are scaling parameters. Consequently, $s_t$ is of dimension $2 + 4 N_{\text{obstacle}}$. The values of all general parameters can be found in Table \ref{tab:EnvsParameters}. We distinguish between an MDP case where the agent can observe the full state at time step $t$:
\begin{equation}
	o_{t,\rm MDP} = s_t,
\end{equation}
and an RV case where the agent only receives positional information about the obstacles:
\begin{equation}
	o_{t,\rm RV} = \begin{pmatrix}
		\frac{\Ddot{y}_{t,\rm agent}}{a_{y,\rm max}}\\
		\frac{\Dot{y}_{t,\rm agent}}{v_{y,\rm max}}\\
		\frac{x_{t,\rm agent}-x_{t,i}}{x_{\rm scale}}\\
		\frac{y_{t,\rm agent}-y_{t,i}}{y_{\rm scale}}
	\end{pmatrix}.
\end{equation}

Based on the observation $o_t$, the agent computes an action $a_t \in [-1,1]$ that is mapped to an acceleration in lateral direction:
\begin{equation}
	\Ddot{y}_{t+1,\rm agent} = \Ddot{y}_{t,\rm agent}  + \Delta a_{y,\rm max} a_t,
\end{equation}
where $\Delta a_{y,\rm max}$ defines the maximal incremental lateral acceleration for the agent. The action can be seen as a jerk and $\Delta a_{y,\rm max}$ as the maximum jerk to avoid too large jumps in the acceleration of the agent. The Euler and ballistic methods are used to update the agent's lateral speed and the positions for agent and obstacles at time step $t+1$ \citep{Treiber2013}. Exemplary for the agent, we have:
\begin{align}
	\Dot{y}_{t+1, \rm agent} &= \Dot{y}_{t, \rm agent} +\Ddot{y}_{t+1, \rm agent}\Delta t,\\
	x_{t+1, \rm agent} &= x_{t, \rm agent} + \frac{\Dot{x}_{t, \rm agent} + \Dot{x}_{t+1, \rm agent}}{2} \Delta t,\\
	y_{t+1, \rm agent} &= y_{t, \rm agent} + \frac{\Dot{y}_{t, \rm agent} + \Dot{y}_{t+1, \rm agent}}{2} \Delta t,
\end{align}
with $\Delta t$ corresponding to the simulation step size.

\begin{table}
\caption{General parameters for the Environments Simple-OA and Complex-OA}
\label{tab:EnvsParameters} 
\begin{center}
	\begin{tabular}{ p{0.15\linewidth} |p{0.5\linewidth} |p{0.15\linewidth} } 
		Parameter & Description & Value   \\ \hline
		$\Delta a_{y,\rm max}$  & agents maximum lateral acceleration change   	& $\unit[0.005]{m/s^2}$   \\
		$a_{y,\rm max}$ 		& agents maximum lateral acceleration  			& $\unit[0.01]{m/s^2}$  \\
		$v_{x,\rm max}$ 		& agents maximum longitudinal speed  	    	& $\unit[5]{m/s}$   \\
		$v_{y,\rm max}$ 		& agents maximum lateral speed 			    	& $\unit[5]{m/s}$    \\
		$\Delta t$              & simulation step size                          & $\unit[5]{s}$
	\end{tabular}
\end{center}
\end{table}

\subsection{Environment Simple-OA}
\begin{figure}[ht]
	\centering
	\includegraphics[width=0.6\linewidth]{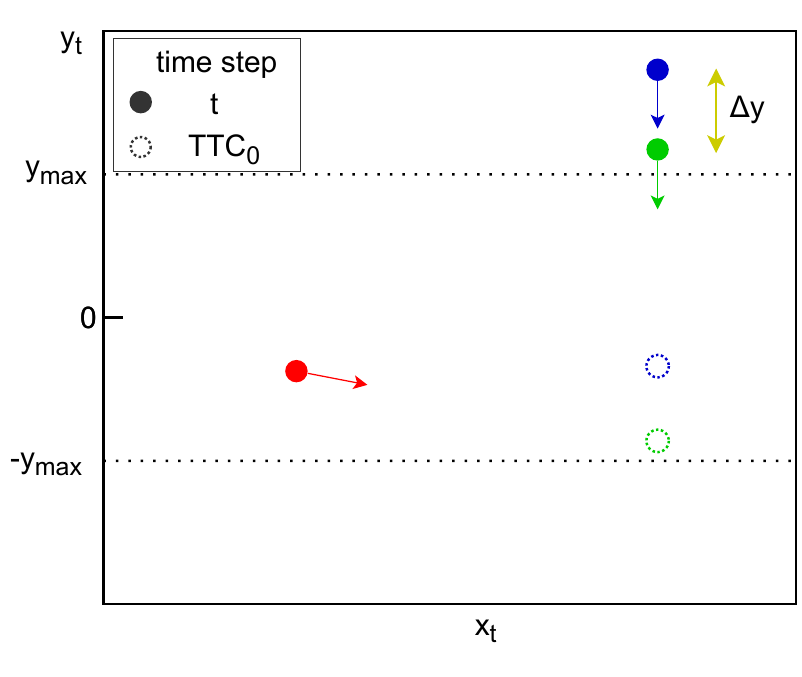}
	\caption{Environment Simple-OA with agent in red and obstacles in blue and green, respectively. The arrows indicate velocities.}
	\label{fig:EnvA_technical}
\end{figure}

The focus of the environment Simple-OA is the isolated analysis of the anticipation of a single trajectory, leading to the specification $N_{\text{obstacle}} = 2$. The obstacles move with the same speed in the lateral direction, while the longitudinal speed is zero. Both obstacles are initialized with the same longitudinal position and the RL agent needs to pass between both obstacles, which can be interpreted as a moving finish line. Figure \ref{fig:EnvA_technical} shows a schematic representation of the environment. The optimal solution of this problem requires to simply anticipate the trajectory of the obstacle pair that moves with a constant lateral speed.

We initialize the state space as follows: The agent's dynamics are zero, except the longitudinal speed $\Dot{x}_{0,agent}$, that is sampled uniformly at random from the interval $[\unit[1]{m/s}, v_{x,\rm max}]$. Further, we sample the agent's initial time-to-collision with the obstacles in longitudinal direction, $TTC_{0}$, uniformly at random from the interval $[\unit[280]{s}, \unit[320]{s}]$. Afterward, the initial dynamics of the two obstacles are set to fulfill the following constraints:
\begin{align}
	\Dot{y}_{t,1} &\sim \mathcal{U}([-v_{y,\rm max}, v_{y,\rm max}]), \\ 
	\Dot{y}_{t,1} &= \Dot{y}_{t,2},\\
	\frac{y_{TTC_{0},1}-y_{TTC_{0},2}}{2} &\sim \mathcal{U}([-y_{\rm max}, y_{\rm max}]),\\
	y_{t,1} - y_{t,2} &= \Delta y,\\
	\Dot{x}_{t,i} &= 0,
\end{align}  
where the parameters $y_{\rm max}$ and $\Delta y$ are described in Table \ref{tab:EnvAParameters} and visualized in Figure \ref{fig:EnvA_technical}.

During one episode, all velocities are kept constant, and an episode ends when:
\begin{equation}
	x_{t,\rm agent} > x_{t,i}.
\end{equation}
The evaluation quantity of interest is whether the final position of the agent is between the obstacles, thus checking whether the trajectory was adequately anticipated. Consequently, we impose a non-zero reward only at the final step of an episode, leading to the following reward structure:
\begin{equation}
	r_t = 
	\begin{cases}
		\phantom{-}100, & \text{if } y_{t,\rm agent} \in (y_{t,2}, y_{t,1}) \text{ and }x_{t,\rm agent} > x_{t,i},\\
		-100, & \text{if } y_{t,\rm agent} \not\in (y_{t,2}, y_{t,1}) \text{ and }x_{t,\rm agent} > x_{t,i},\\
			\phantom{-10}0, & \text{otherwise}.
	\end{cases}
\end{equation}

\begin{table}
	\caption{Parameters for the environment Simple-OA}
	\label{tab:EnvAParameters} 
	\begin{center}
		\begin{tabular}{ p{0.15\linewidth} |p{0.5\linewidth} |p{0.15\linewidth} } 
			Parameter 	& Description & Value   \\ \hline
			$N_{\text{obstacle}}$ & number of obstacles & 2 \\
			$x_{\rm scale}$	& scaling factor for observation  	& $\unit[1500]{m}$   \\
			$y_{\rm scale}$ & scaling factor for observation  	& $\unit[1700]{m}$   \\
			$y_{\rm max}$ 	& end zone for obstacles		  	& $\unit[200]{m}$   \\
			$\Delta y$ 	& distance between obstacles	  	& $\unit[50]{m}$   				
		\end{tabular}
	\end{center}
\end{table}

\subsection{Environment Complex-OA}
The environment Complex-OA represents a more realistic environment for obstacle avoidance. In contrast to the environment Simple-OA, where the agent had to anticipate a single obstacle trajectory, the agent now has to anticipate several trajectories simultaneously. A further challenge is to prioritize those trajectories regarding their potential of leading to a collision in the near future.

\begin{figure}[ht]
	\centering
	\includegraphics[width=0.85\linewidth]{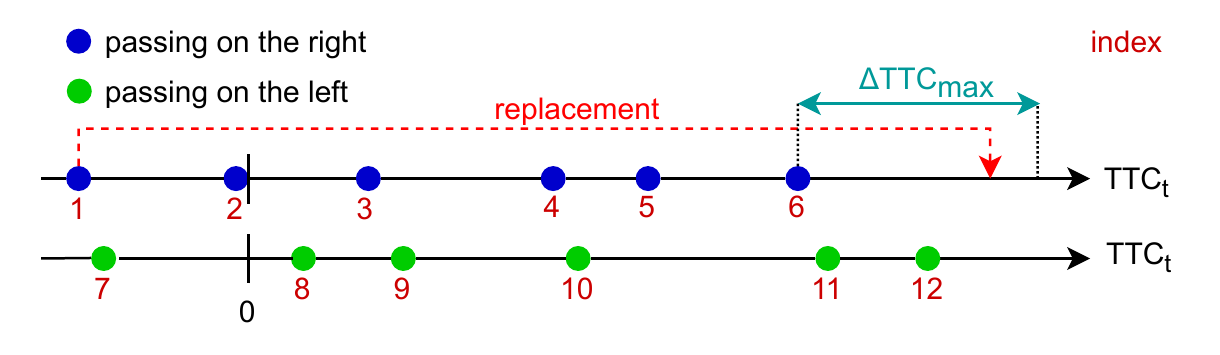}
	\caption{Replacement of an obstacle (obstacle $1$) since two obstacles with the same passing rule already passed the agent (negative time-to-collision). The obstacle's new $TTC_{t,1}$ is set uniformly at random in the time interval colored turquoise with the length of $\Delta TTC_{\rm max}$.}
	\label{fig:EnvB_TTC}
\end{figure}

We define our observation space with $N_{\text{obstacle}} = 12$. Further, we define the set of obstacles that should only be passed, from the perspective of the agent, on the right side in lateral direction as $\mathcal{M}_{\rm right} = \{1, \ldots, N_{\text{obstacle}} /2\}$. Consequently, the remaining obstacles should be passed left and are denoted $\mathcal{M}_{\rm left} = \{N_{\text{obstacle}}/2 + 1, \ldots, N_{\text{obstacle}}\}$. Similar to the Simple-OA, we initialize the agent's dynamics to zero, except the longitudinal speed $\Dot{x}_{0,\rm agent}$, that is sampled uniformly at random from the interval $[\unit[1]{m/s},v_{x,\rm max}]$. We define $TTC_{t,i}$ as the agent's time-to-collision with an obstacle $i \in \mathcal{M}$ in longitudinal direction at time step $t$. Negative values for $TTC_{t,i}$ relate to obstacles that already passed the agent in the longitudinal direction. If for two obstacles $k, l \in \mathcal{M_{\rm right}}$ holds: $TTC_{t,k} < 0$, $TTC_{t,l} < 0$, and $TTC_{t,k} < TTC_{t,l}$, we replace obstacle $k$ as shown in Figure \ref{fig:EnvB_TTC}. Its new time-to-collision is randomly sampled from:
\begin{equation}
	\label{eq:EnvBTTC}
TTC_{t,k} \sim \mathcal{U}\left(\left[\max_{j \in \mathcal{M_{\rm right}}}(TTC_{t,j}), \max_{j \in \mathcal{M_{\rm right}}}(TTC_{t,j}) + \Delta TTC_{\rm max}\right]\right),
\end{equation}
where $\Delta TTC_{\rm max}$ is the maximal temporal distance for the new placement of an obstacle. This parameter affects the number of obstacles being passed in a certain time interval and is therefore a crucial design element of this environment. The same replacement procedure is applied for obstacles with passing rule 'left'.

\begin{figure}[ht]
	\centering
	\includegraphics[width=0.8\linewidth]{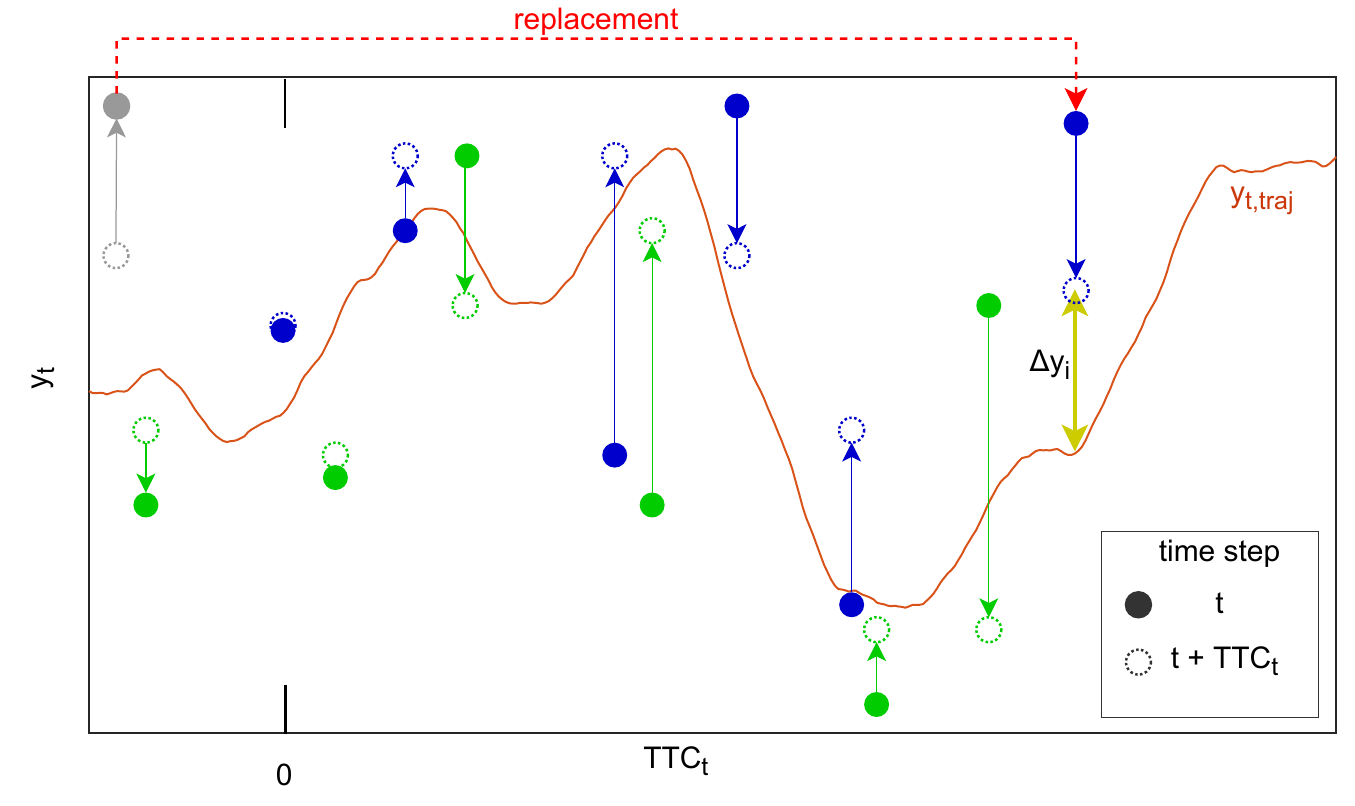}
	\caption{Replacement of an obstacle identical to the situation in Figure \ref{fig:EnvB_TTC} but with additional information about lateral positions of obstacles.}
	\label{fig:EnvB_technical}
\end{figure}

Having computed the new value for $TTC_{t,i}$ for a replaced obstacle $i \in \mathcal{M}$ at time step $t$, the new dynamics of the obstacle need to be set. First, we draw values for $\Dot{x}_{t,i}$ and $\Dot{y}_{t,i}$ from uniform distributions:
\vspace{-0.5cm}
\begin{align}	
	\Dot{x}_{t,i} &\sim \mathcal{U}([-v_{x,\rm max}, v_{x,\rm max}]), \label{eq:EnvBxDot} \\		
	\Dot{y}_{t,i} &\sim \mathcal{U}([-v_{y,\rm max}, v_{y,\rm max}]).
	\label{eq:EnvByDot}
\end{align} 
Second, the new longitudinal position can be set according to:
\begin{equation}
	\label{eq:EnvBx}
	x_{t,i} = (\Dot{x}_{t,\rm agent} - \Dot{x}_{t,i})  TTC_{t,i} + x_{t,\rm agent}.
\end{equation}
Third, having the lateral speed of the replaced obstacle set, we generate the new lateral position with the help of a predefined, stochastic trajectory $y_{t,\rm traj}$. This lateral trajectory is computed at the beginning of an episode and is based on a smoothed AR(1) process \citep{tsay2005analysis}, whose parameters reflect the kinematics of the agent. Figure \ref{fig:EnvB_technical} shows a replacement situation identical to Figure \ref{fig:EnvB_TTC} and illustrates how this trajectory is used to define the new lateral position for a replaced obstacle. One can think of this stochastic process as an approximate trajectory the agent has to follow to avoid collisions with obstacles. In the following, we define the smoothed AR(1) process and give a detailed explanation about the replacement of an obstacle based on that process.

The AR(1) process is defined as:
\begin{equation} 
	X_{t+1}=\phi X_t+u, \quad \text{where} \quad u \sim \mathcal{N}(0, \sigma_{\rm AR}^2),
\end{equation}
with auto-regressive parameter $\phi$ and variance $\sigma_{\rm AR}$. The parameters have been designed to model a lateral trajectory the agent can approximately follow under acceleration and velocity constraints represented by $a_{y,\rm max}$ and $v_{y,\rm max}$. To reduce the noise, we exponentially smooth the AR(1) process:
\vspace{-0.3cm}
\begin{align}
	&y_{t,\rm traj}=
	\begin{cases}
		X_{0}, & \text{for } t = 0,\\
		\beta X_{t}+(1-\beta) y_{t-1,\rm traj}, & t >0,
	\end{cases} 
\end{align}
where $\beta$ defines the smoothing factor. Based on this trajectory and having already computed $TTC_{t,i}$, $x_{t,i}$, $\Dot{x}_{t,i}$, and $\Dot{y}_{t,i}$ via (\ref{eq:EnvBTTC}), (\ref{eq:EnvBxDot}), (\ref{eq:EnvByDot}), and (\ref{eq:EnvBx}), one more step is needed to set the new lateral position $y_{t,i}$ for a replaced obstacle $i \in \mathcal{M}$ at time step $t$. 

We define $\Delta y_{i}$ as the absolute difference between an obstacle's lateral position $y_{t,i}$ and the defined trajectory $y_{t,\rm traj}$ when agent and obstacle are at the same longitudinal position ($TTC_{t,i} = 0$):
\begin{equation}
\Delta y_{i} = |y_{t,i} - y_{t,\rm traj}| \quad \text{for} \quad TTC_{t,i} = 0,
\end{equation}
shown yellow in Figure \ref{fig:EnvB_technical}.
To force our agent to move approximately along the trajectory $y_{t,\rm traj}$, the positional difference $\Delta y_{i}$ should be small, thus being another crucial design parameter to adjust the complexity of the environment. Every time an obstacle $i$ is replaced, the variable $\Delta y_{i}$ is sampled from a normal distribution:
\begin{equation}
\Delta y_{i}  \sim  \mathcal{N}(\mu_{\Delta y},\sigma_{\Delta y}^2),
\end{equation} 
and lower-bounded to $\Delta y_{\rm min}$:
\begin{equation}
\Delta y_{i} = \max(\Delta y_{\rm min}, \Delta y_{i}).		
\end{equation}
By changing the parameters $\Delta y_{\rm min}$, $\sigma_{\Delta y}^2$, and $\Delta y_{\rm min}$, one can adjust how close the obstacles are coming to the trajectory $y_{t,\rm traj}$ when obstacle and agent are at the same longitudinal position. The chosen values for those parameters can be found in Table \ref{tab:EnvBParameters}. Finally, the lateral position for obstacles $i_R \in \mathcal{M_{\rm right}}$ is computed via:
\begin{equation}
y_{t,i_R} = y_{t,\rm traj} + \Delta y_{i_R} - \Dot{y}_{t,i_R}TTC_{t,i_R},
\end{equation}
and for obstacles $i_L \in \mathcal{M_{\rm left}}$ via:
\begin{equation}
y_{t,i_L} = y_{t,\rm traj} - \Delta y_{i_L} - \Dot{y}_{t,i_L}TTC_{t,i_L}.
\end{equation}
Figure \ref{fig:EnvB_technical} shows the final lateral position and time-to-collision for a replaced obstacle as a filled circle.

\begin{figure}[ht]
	\centering
	\includegraphics[width=0.85\textwidth]{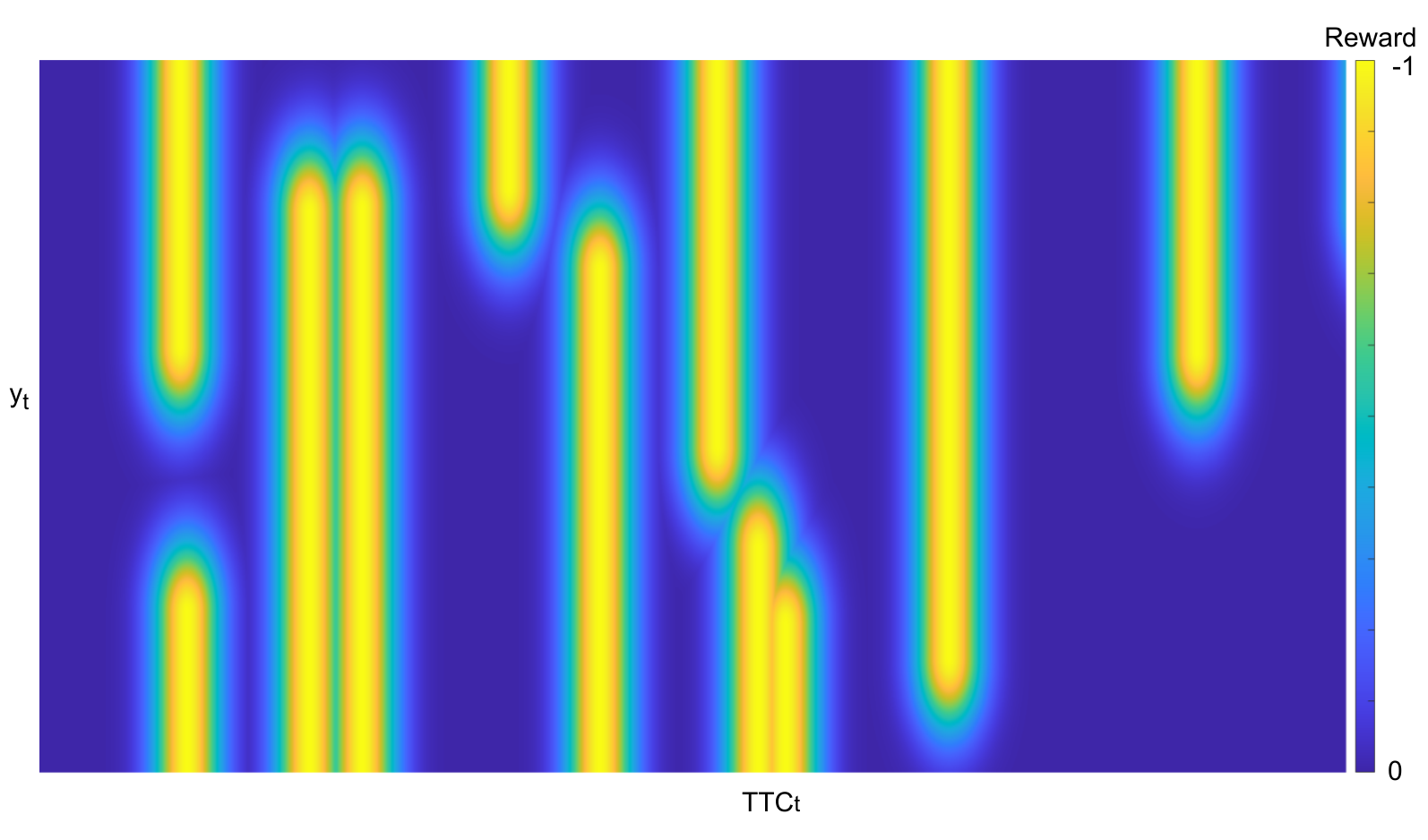}
	\caption{Reward function for the environment Complex-OA.}
	\label{fig:reward}
\end{figure}

In the following, we detail the reward function used to train the RL agent. The aim of this function is to penalize collisions with obstacles, to consider the passing rule for each obstacle, and to penalize getting in the proximity of an obstacle. Considering all these factors, we define the reward for an obstacle $i_R \in \mathcal{M_{\rm right}}$ at time step $t$ as:
\begin{equation}
    r_{t,i_R} =  -\frac{\varphi( TTC_{t,i_R} / \sigma_{TTC})}{\varphi(0)} \frac{\varphi\{\max(0, y_{t,i_R} - y_{t,\rm agent})/\sigma_y\}}{\varphi(0)}, 
\end{equation}
and for an obstacle $i_L \in \mathcal{M_{\rm left}}$:
\begin{equation}
    r_{t,i_L} = -\frac{\varphi( TTC_{t,i_L} / \sigma_{TTC})}{\varphi(0)}\frac{\varphi\{\max(0,  y_{t,\rm agent}-y_{t,i_L})/\sigma_y\}}{\varphi(0)} ,
\end{equation}
where $\varphi(x)$ denotes the density function of the standard normal distribution and the parameters $\sigma_{TTC}^2$ and $\sigma_y^2$ describe variances. Since we are only interested in penalizing the agent with respect to the closest obstacle, the reward at time step $t$ is defined to be the minimum of all obstacle rewards:
\begin{equation}
    r_t = \min_{i \in \mathcal{M}}(r_{t,i}).
\end{equation}
Figure \ref{fig:reward} illustrates the reward function for twelve obstacles. As one can see, violating the passing rule is penalized in the same magnitude as colliding with an obstacle. Furthermore, the agent is also penalized when getting close to an obstacle, adjusted by the parameters  $\sigma_{TTC}^2$ and $\sigma_y^2$. At this point it is important to mention that when the agent passes obstacles with different relative longitudinal speeds, the agent should be rewarded in the same way. Therefore, we use the time-to-collision instead of the position in longitudinal direction to model the reward function. 

Table \ref{tab:EnvBParameters} contains a description and the chosen values for all Complex-OA parameters.

\begin{table}[H]
\caption{Parameters for the environment Complex-OA}
\label{tab:EnvBParameters} 
\begin{center}
	\begin{tabular}{ p{0.15\linewidth} |p{0.6\linewidth} |p{0.15\linewidth} } 
	Parameter & Description & Value   \\ \hline
	$N_{\text{obstacle}}$ & number of obstacles & 12 \\
		$x_{scale}$			& scaling factor for observation  					& $\unit[3000]{m}$   \\
		$y_{scale}$ 		& scaling factor for observation  					& $\unit[3000]{m}$   \\
		$\Delta TTC_{max}$	& maximal temporal distance for replacing an obstacle	& $\unit[300]{s}$   \\
		$\phi$ 		        & AR(1) process	parameter		                	& $0.99$  \\
        $\sigma_{\rm AR}$ 	& normal distribution variance      		    	& $\unit[28.3]{m^2}$\\
		$\beta$ 			& smoothing factor 									& $0.03$   \\
		$\mu_{\Delta y}$ 	& normal distribution mean							& $\unit[100]{m}$   \\
		$\sigma_{\Delta y}^2 $ & normal distribution variance					& $\unit[50]{m^2}$   \\
		$\Delta y_{\rm min}$& minimum bound for $\Delta y$						& $\unit[40]{m}$   \\
		$\sigma_{y}^2 $     & normal distribution variance	    				& $\unit[25]{m^2}$   \\
		$\sigma_{TTC}^2 $   & normal distribution variance				    	& $\unit[25]{s^2}$   \\

\end{tabular}
\end{center}
\end{table}

\section{Results}\label{sec:results}
We train the TD3, LSTM-TD3, and TD3-FS algorithms for both environments, Simple-OA and Complex-OA. The frame-stacking consists of expanding the current observation with the observations from the last two steps to match the LSTM-TD3 information set with $l = 2$. The training setup and hyperparametrization is identical to Section \ref{subsec:implementation}, except that we train for $15\cdot 10^6$ time steps since we could not observe convergence beforehand. Regarding the Simple-OA, the return yields a straightforward interpretation since, e.g., a return of 80 implies that 9 out of the 10 evaluation episodes have been successful. In the following, we summarize the main findings of this investigation:

\begin{enumerate}
    \item For both environments, the algorithms perform worse if there is no velocity information.
    \item In the environment Simple-OA, the TD3 fails nearly completely if no velocity information is available since the agent rarely passes between the obstacles. In the contrary, LSTM-TD3 and TD3-FS are significantly better than TD3 and perform on a comparable level, although still not reaching the MDP performance.
    \item In the environment Simple-OA, the LSTM-TD3 algorithm learns a near optimal policy in a fraction of the considered training steps and overall dramatically stabilizes the learning process compared to the TD3 and TD3-FS approaches for the MDP scenario.
    \item In the environment Complex-OA, all algorithms perform similarly in both observation space configurations. However, the TD3-FS appears slightly worse than its two competitors.
\end{enumerate}

\begin{figure}[H]
    \centering
    \includegraphics[width=\textwidth]{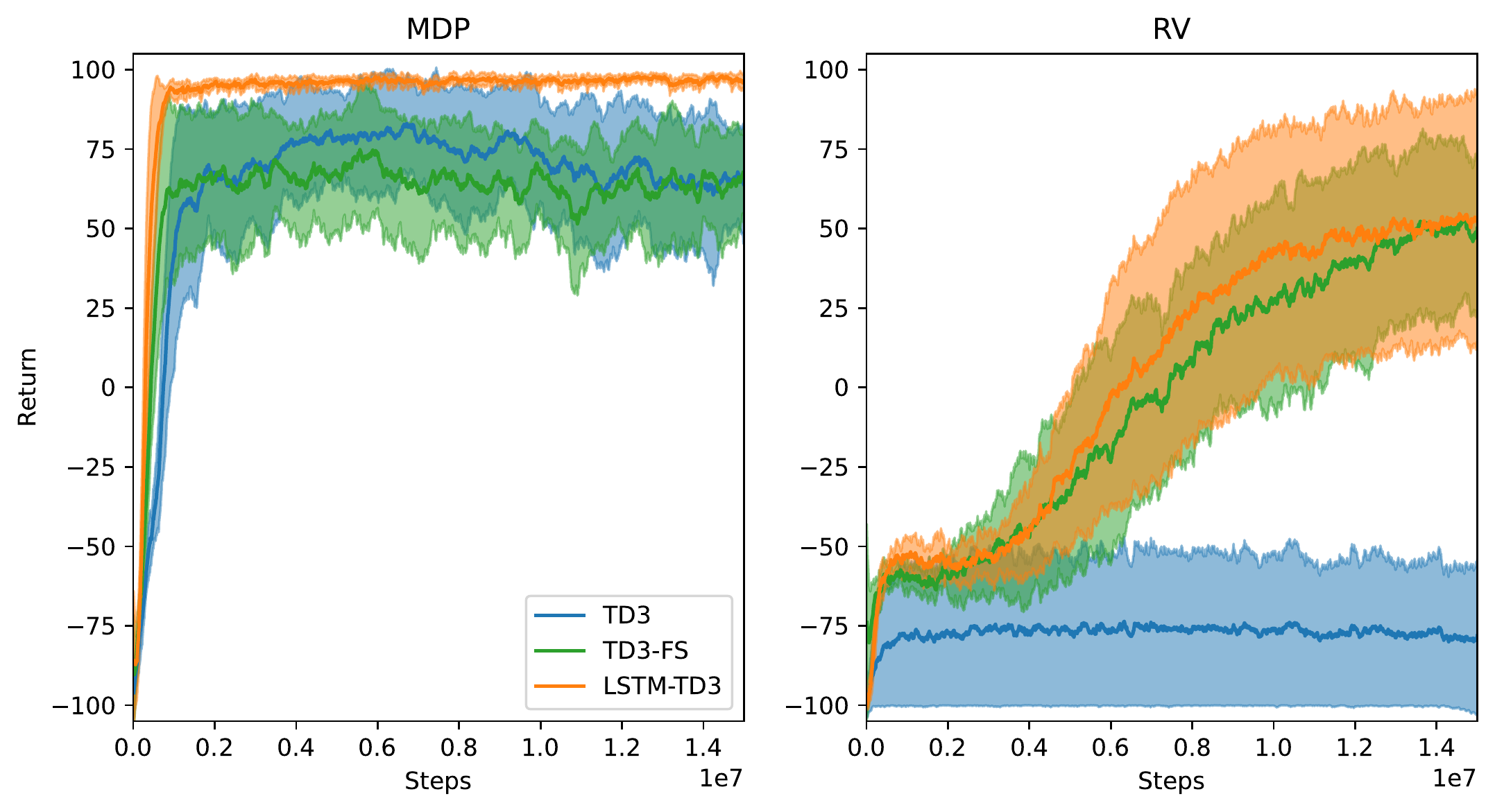}
    \caption{Performance comparison of the considered agents in the environment Simple-OA. Results are averaged over 10 independent runs. The shaded area are two standard deviations over the runs.}
    \label{fig:Ski_final_results}
\end{figure}

\begin{figure}[H]
    \centering
    \includegraphics[width=\textwidth]{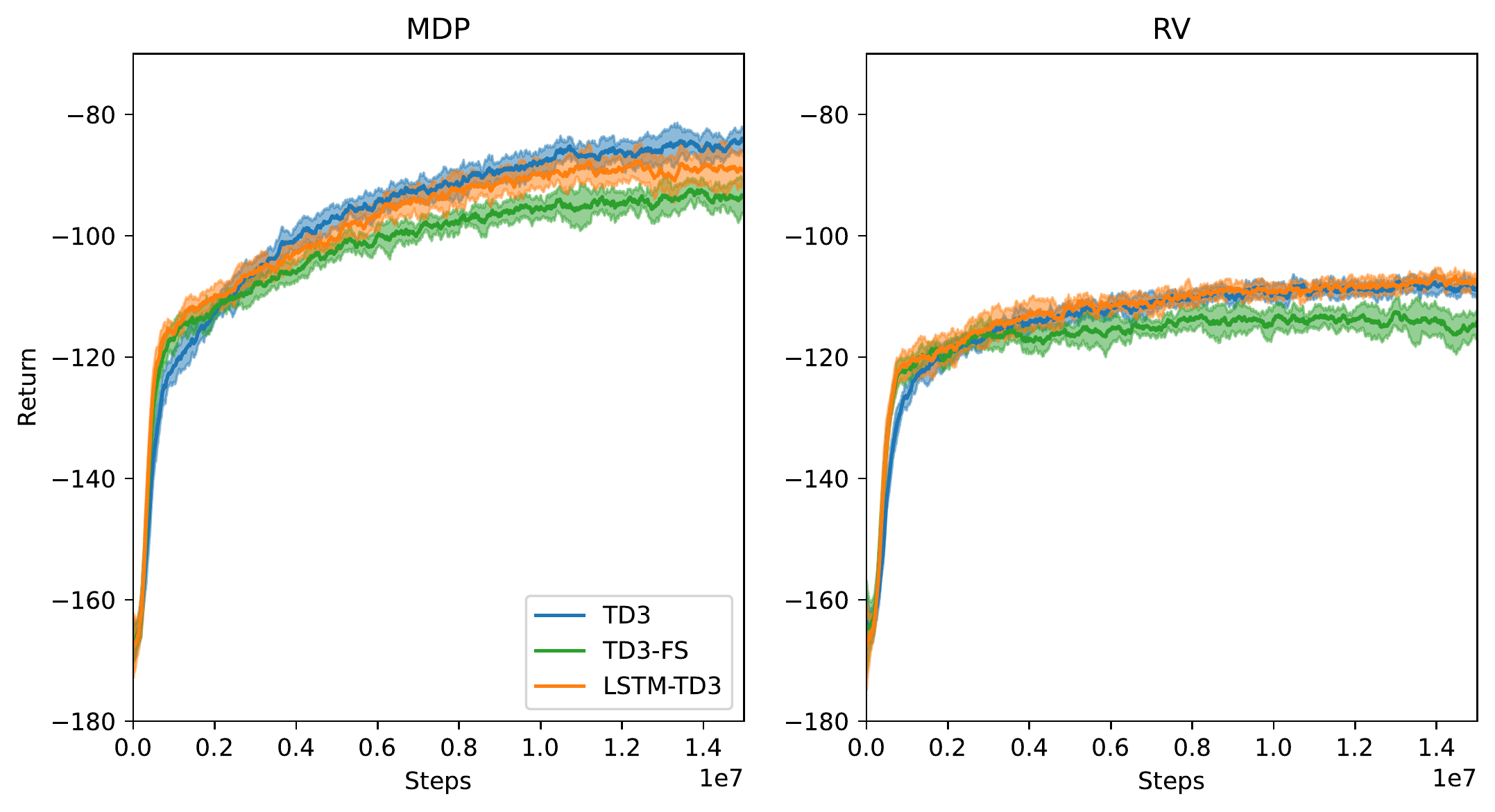}
    \caption{Performance comparison of the considered agents in the environment Complex-OA. Results are averaged over 10 independent runs. The shaded area are two standard deviations over the runs.}
    \label{fig:OA_final_results}
\end{figure}

\section{Discussion}
\label{sec:Discussion}
In the environment Simple-OA, the optimal policy simply requires anticipating one particular trajectory of a randomly generated obstacle pair, representing a basic analytical task. Astonishingly, even with all necessary information, both agents without recurrency in their respective function approximators cannot master this task entirely. Similarly surprising, although the recurrent and frame-stacking approaches perform significantly better than a 'plain' agent in the RV scenario, they are even in this simple task not able to achieve a similar return level as in the MDP case.

In contrast to the environment Simple-OA, the Complex-OA scenario yields two additional difficulties. First, several trajectories need to be simultaneously anticipated. Second, this information needs to be processed by prioritizing the trajectories regarding their potential of generating a collision in the near future. All agents perform reliably better when provided with the complete state information for this environment while displaying a performance drop in the RV scenario. This is in line with the findings from the Simple-OA. Remarkably, in contrast to the first task, the recurrency and frame-stacking approaches do not help to solve the Complex-OA. More precisely, the recurrent agent learns a performance-wise equal policy like the 'plain' agent. This leads to the assumption that the recurrent agent also solely relies on positional information without developing the ability to anticipate and prioritize the obstacles' trajectories correctly. We argue that this can be explained by the increased complexity of the environmental dynamics contrary to the Simple-OA scenario, which is an isolated investigation concerning only one relevant trajectory. To guarantee the robustness of these findings against different specifications of the environments, we tested for a variety of different settings by changing: the number of obstacles, the maximum temporal distance for replacing an obstacle, the smoothed AR(1) process parametrization, the obstacle distance to the AR(1)-based trajectory when longitudinal TTC is zero, the maximum acceleration and velocities for agent and obstacles, the reward configuration, and the RL algorithm hyperparameters. Throughout all these specifications, the main findings are qualitatively unchanged.

Regarding our initial Hypothesis 1 \& 2, we can reject both of them. The recurrent layers as well as frame-stacking approaches are not able to consistently replace missing velocity information in the observation space. However, in simplified scenarios, they are able to significantly boost performance and stabilize the overall training procedure.

\section{Conclusion}
\label{sec:Conclusion}
Dynamic obstacle avoidance is a fundamental task in many real-world application domains, e.g., self-driving cars, service robots, or unmanned surface vehicles. A core element of successfully mastering obstacle avoidance tasks is the precise anticipation of trajectories of relevant obstacles. However, real-world systems rely on sensor data that is often limited to positional information about moving obstacles or vehicles without explicitly providing velocity-related information. From a physical perspective, trajectories cannot be predicted solely from current positional information, and approaches like recurrency or frame-stacking are expected to yield improved performance. We analyze the severity of missing velocity information and evaluate the potential of recurrency and frame-stacking approaches. Therefore, we define a traffic type independent environment with variants of different complexity, in which we test several model-free RL agents. Across all agents, we found that the lack of velocity information significantly harms the performance. The approaches of recurrency and frame-stacking cannot reliably replace missing velocity information in the observation space. In complex dynamic obstacle avoidance scenarios, which require anticipating and prioritizing the trajectories of several objects, all agents struggle to reach the performance they achieve with complete information. However, in simplified scenarios, where the anticipation of a single trajectory is required, recurrency and frame-stacking can significantly improve the agent's ability to learn an appropriate obstacle avoidance behavior even when only positional information is available. Although it does not always improve the agent's performance, we generally recommend to integrate recurrency in the function approximators when only positional sensor data is available.

\subsubsection*{Acknowledgements}
This work was partially funded by BAW - Bundesanstalt für Wasserbau (Mikrosimulation des Schiffsverkehrs auf dem Niederrhein).

\bibliographystyle{asa}
\bibliography{bib}

\newpage
\appendix
\section{Algorithm details and hyperparameter}\label{appendix:TD3_algo}

\begin{algorithm}[h]
\setstretch{1.05}
\SetAlgoLined
 Randomly initialize critics $Q^{\omega_1}, Q^{\omega_2}$ and actor $\mu^{\theta}$\\
 Initialize target critics $Q^{\omega'_1}, Q^{\omega'_2}$ and target actor $\mu^{\theta'}$ with $\omega'_1 \leftarrow \omega_1$, $\omega'_2 \leftarrow \omega_2$, $\theta^{'} \leftarrow \theta$\\
 Initialize replay buffer $\mathcal{D}$\\
 Receive initial state $s_1$ from environment \\
 \For{t = 1,T}{
 \emph{Acting}\\
 Select action with exploration noise: $a_t = \mu^{\theta}(s_t) + \epsilon$, \quad $\epsilon \sim \mathcal{N}(0, \sigma)$\\
 Execute $a_t$, receive reward $r_{t+1}$, new state $s_{t+1}$, and done flag $d_t$\\
 Store transition $(s_t, a_t, r_{t+1}, s_{t+1}, d_t)$ to $\mathcal{D}$\\
 \medskip
 \emph{Learning}\\
  Sample random mini-batch of transitions $(s_i, a_i, r_{i+1}, s_{i+1}, d_i)_{i=1}^{N} $ from $\mathcal{D}$\\
  Calculate targets:\\
  \vspace{-1cm}
  \begin{align*}
      \Tilde{a}_{i+1} &= \mu^{\theta'}(s_{i+1}) + \Tilde{\epsilon}, \quad \Tilde{\epsilon} \sim \text{clip}\{\mathcal{N}(0, \Tilde{\sigma}),-c,c\},\\
      y_i &= r_{i+1} + \gamma (1-d_i) \min_{j=1,2} Q^{\omega'_j}(s_{i+1},\Tilde{a}_{i+1}).
  \end{align*} \\
  \vspace{-0.5cm}
 Update critics: $\omega_j \leftarrow \min_{\omega_j}N^{-1} \sum_i \left\{y_i - Q^{\omega_j}(s_i, a_i)\right\}^2$\\
 \If{$t \mod d$}{
 Update actor: $\theta \leftarrow \max_{\theta} N^{-1} \sum_{i} Q^{\omega_1}\left\{s_i, \mu^{\theta}(s_i)\right\}$\\
  Update target networks via (\ref{eq:DDPG_soft_tgt_up})
 }
 \medskip
\emph{End of episode handling}\\
    \If{$d_t$}{
    Reset environment to an initial state $s_{t+1}$\\
    }
}
 \caption{TD3 algorithm following \cite{fujimoto2018addressing}.}
 \label{algo:TD3}
\end{algorithm}

\begin{table}[H]
    \centering
    \begin{tabular}{l|l}
    Hyperparameter & Value\\
    \toprule
        Discount factor $\gamma$ &  0.99 \\
        Batch size $N$ & 32\\
        Replay buffer size $|\mathcal{D}|$ & $10^5$ \\
        Learning rate actor $\alpha_{actor}$ & $10^{-4}$ \\ 
        Learning rate critic $\alpha_{critic}$ & $10^{-4}$ \\ 
        Target update rate $\tau$ & 0.001 \\
        Random start step $N_{start\_step}$ & $5\,000$ \\
        Update start step $N_{update\_after}$ & $5\,000$ \\
        Optimizer & Adam \\
        Exploration noise $\sigma$ & 0.1 \\
        Target policy smoothing $\Tilde{\sigma}$ & 0.2\\
        Target policy smoothing $c$ & 0.5\\
        Policy update delay $d$ & 2\\
        History length $l$ & 2 \\
    \end{tabular}
    \caption{List of hyperparameters used in both TD3 and LSTM-TD3. $N_{start\_step}$ means that at the beginning of each training process, the agent performs $N_{start\_step}$ steps completely at random for initial exploration. $N_{update\_after}$ is the step after which critic and actor updates are performed.}
    \label{tab:hyperparams}
\end{table}

\end{document}